\documentclass[lettersize,journal]{IEEEtran}
\usepackage{amsmath,amsfonts}
\usepackage{algorithmic}
\usepackage{algorithm}
\usepackage{array}
\usepackage[caption=false,font=normalsize,labelfont=sf,textfont=sf]{subfig}
\usepackage{textcomp}
\usepackage{stfloats}
\usepackage{url}
\usepackage{verbatim}
\usepackage{graphicx}
\usepackage{cite}
\usepackage{utfsym,colortbl,overpic}
\begin{document}

\title{TSSR: A Truncated and Signed Square Root Activation Function for Neural Networks}

\author{Yuanhao Gong\\College of Electronics and Information Engineering, Shenzhen University, China.~~gong.ai@qq.com}

\markboth{Journal of \LaTeX\ Class Files,~Vol.~14, No.~8, August~2021}%
{Shell \MakeLowercase{\textit{et al.}}: A Sample Article Using IEEEtran.cls for IEEE Journals}


\maketitle

\begin{abstract}
Activation functions are essential components of neural networks. In this paper, we introduce a new activation function called the Truncated and Signed Square Root (TSSR) function. This function is distinctive because it is odd, nonlinear, monotone and differentiable. Its gradient is continuous and always positive. Thanks to these properties, it has the potential to improve the numerical stability of neural networks. Several experiments confirm that the proposed TSSR has better performance than other stat-of-the-art activation functions. The proposed function has significant implications for the development of neural network models and can be applied to a wide range of applications in fields such as computer vision, natural language processing, and speech recognition.
\end{abstract}


\section{Introduction}
\IEEEPARstart{A}{ctivation} functions are a crucial component of neural networks, playing a critical role in determining their success. These functions introduce non-linearity to the network, which is essential. Without non-linearity, the neural network would just be a linear regression model, limiting its power.

Activation functions are applied to the output of nodes in a neural network to determine whether the node should be activated based on its input. This is done by applying a mathematical function to the input of the node, and then passing the output on to the next layer of nodes in the network.

The three most commonly used activation functions are sigmoid, ReLU, and tanh. Sigmoid is a smooth function that maps any input to a value between 0 and 1, making it useful for binary classification tasks. ReLU is a non-linear function that returns the input if it is positive and 0 if it is negative, making it useful for rectifying the output of nodes in the network. Tanh is a scaled version of the sigmoid function that maps inputs to a value between -1 and 1, making it useful for tasks where both positive and negative values are important.

Selecting the right activation function is critical, as it significantly impacts the performance of the network. The choice of activation function depends on the specific task that the neural network is being used for and the architecture of the network. It's essential to pay close attention to the task and carefully evaluate the network before selecting the right activation function to ensure optimal performance.

\begin{figure}
	\centering
		\begin{overpic}[width=0.45\linewidth]{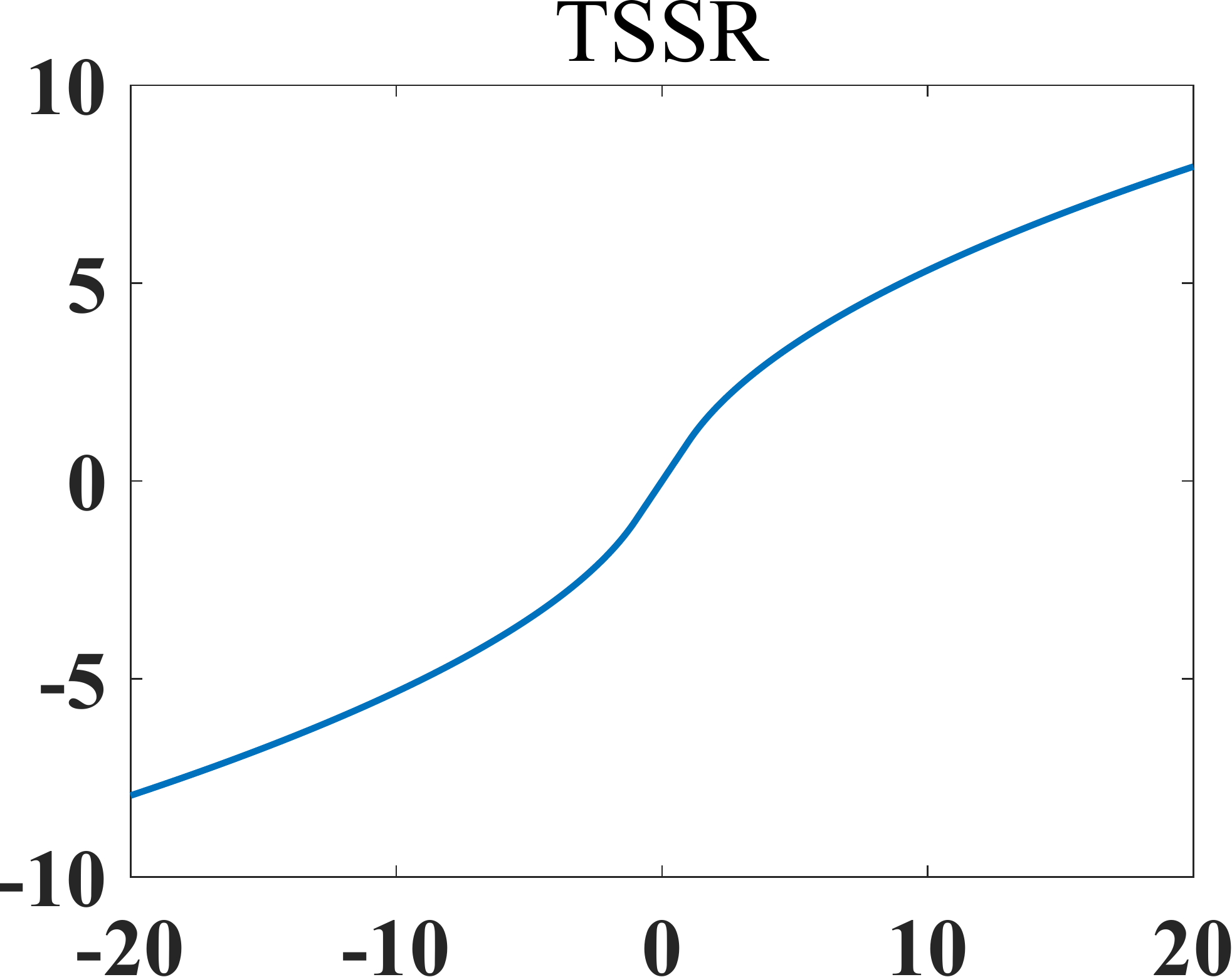}
			\put(20,13){$\footnotesize {f=\left\{
				\begin{array}{l}
					x,\mathrm{if}~ |x|\le 1 \\
					\delta(x)(2\sqrt{|x|}-1)
				\end{array}\right.}$}
		\end{overpic}
		~~~
		\begin{overpic}[width=0.45\linewidth]{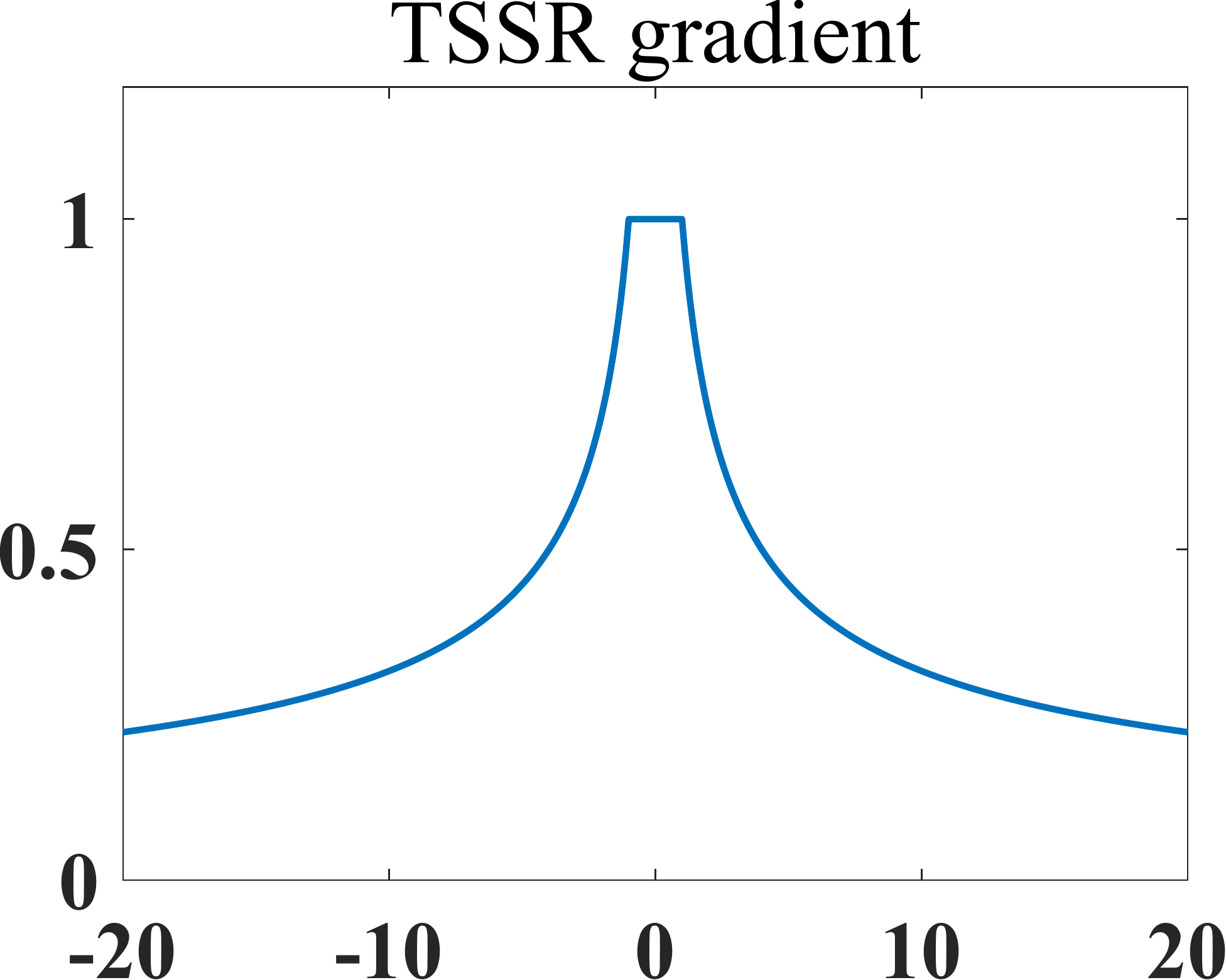}
			\put(20,15){$\footnotesize {f'=\left\{
					\begin{array}{l}
						1,|x|\le 1\\
						|x|^{-\frac{1}{2}}
					\end{array}\right.}$}
		\end{overpic}
	\caption{The proposed activation function (left) and its gradient (right), where $\delta(x)$ is the sign of $x$. This activation function has several desired properties such as being odd, monotone, differentiable and having positive gradient.}
	\label{fig1}
\end{figure}
\subsection{Related Work}
\label{sec:work}
There are some well-known activation functions available. The famous $Sigmoid$ function is defined as
\begin{equation}
	f_1(x)=\frac{1}{1+\exp(-x)}\,,
\end{equation} whose values are bounded in $(0,1)$.
The popular $ReLU$ is defined as
\begin{equation}
	f_2(x)=\max(x,0)\,,
\end{equation} whose values are bounded in $(0,+\infty)$. It has some variants such as $PReLU$, which is defined as
\begin{equation}
	f_3(x)=\left\{
	\begin{array}{ll}
		x, &\mathrm{when }\, x>0 \\
		\alpha x, &\mathrm{else}
	\end{array}\right.,
\end{equation} whose values are not bounded. Another variant $ELU$ is
\begin{equation}
	f_4(x)=\left\{
	\begin{array}{ll}
		x, &\mathrm{when }\, x>0 \\
		\alpha (e^x-1), &\mathrm{else}
	\end{array}\right..
\end{equation} Another activation $swish$ is defined as
\begin{equation}
	f_5(x)=x*Sigmoid(x)\,,
\end{equation} which has a negative lower bound. The $\tanh$ function is 
\begin{equation}
	f_6(x)=\frac{e^{x}-e^{-x}}{e^{x}+e^{-x}}\,,
\end{equation} whose values are in $(-1,1)$. The $softsign$ function is
\begin{equation}
	f_7(x)=\frac{x}{|x|+1}\,,
\end{equation}whose value are also in $(-1,1)$. The well-known $softmax$ is
\begin{equation}
	f_8(x_i)=\frac{e^{x_i}}{\sum_{i}e^{x_i}}\,,
\end{equation} which is frequently used in classification tasks. Very recently, $f_8$ has been shown to lead to numerical issues in~\cite{Bondarenko2023}. An activation function in~\cite{Li2020a} named the Soft-Root-Sign is 
\begin{equation}
	f_9(x)=\frac{x}{\frac{x}{\alpha}+\exp(-\frac{x}{\beta})}\,.
\end{equation} Recently, an activation function named Serf is defined as~\cite{Nag2023}
\begin{equation}
	f_{10}(x)=x*\mathrm{erf}(\log(e^x+1))\,,
\end{equation} where $erf$ is the error function.

Although researchers might design their own activation functions for specific tasks, there are some mathematical properties that activation functions should have to ensure the numerical stability of the neural networks. We will analyze these preferred properties in the following section.
\subsection{Analysis}
As already mentioned, activation functions play a crucial role in providing non-linearity to neural networks. This non-linearity enables neural networks to capture complex patterns and relationships among input data that would otherwise be difficult to identify. In addition to this, activation functions are also responsible for the output range of a neural network, which can have a significant impact on the performance of the network. It is important to choose an appropriate activation function that can effectively handle the specific task at hand and produce accurate results. Therefore, a thorough understanding of the behavior and properties of different activation functions is essential for building and training effective neural networks.

In general, there are two types of activation functions. One is centered at the origin and anti-symmetric. This type of activation function allow negative values in the output and usually are adopted for regression tasks. And the other is $f(x)\ge 0$ that are more suitable for classification tasks, where the output usually indicates category labels. 

A proper activation function has to satisfy several mathematical properties that can help in improving the numerical stability of the neural networks. We analyze some in the following aspects.
\subsubsection{odd function}
We believe that the activation function should be an odd function, $f(-x)=-f(x)$. From mathematical point of view, it is strange to prefer the positive values than the negative values. And such preference might implicitly bias the learning system. (Please distinguish the bias with the bias in the linear transformation). To eliminate such bias, it is better to have an odd function as an activation function. 
\subsubsection{monotone function}
We also believe that the activation function should be monotone and, in most of cases, non-decreasing. Such property preserves the order in the input. In other words, the larger input is non-linearly mapped into a larger output. This order preserving property is desired. And a monotone function usually is a bijective mapping, which means the output does not lose the information from the input. 
\subsubsection{differentiable}
Another property that an activation function should have is differentiable. With such property, the gradient of the activation function is continuous. Thus the gradient function has no dramatic change in a small neighborhood. The continuity of the gradient function guarantees the numerical stability when performing the back-propagation algorithm.
\subsubsection{unbounded value}
The value of an activation function should fully fill the interval $(-\infty,+\infty)$. In contrast, the function with bounded values such as $ReLU$ will have small difference when two inputs have negative values. For example, $ReLU(-1000)= 0= ReLU(-10)$, although $-1000$ and $-10$ have a significant numerical difference. In other words, the $ReLU$ activation function can not distinguish the two input $-1000$ and $-10$, showing its limitations. 

\subsubsection{continuous gradient}
On the other hand, the gradient of the activation function should be continuous and nonzero. The zero gradient (also known as vanishing gradient) is problematic when the back-propagation algorithm is performed. According the monotone property, we expect the gradient is continuous. The continuity guarantees that there is no dramatic change in a small neighborhood. It helps in improving the numerical stability of the neural networks.

We consider these five properties as desired properties of activation functions. And we will evaluate previous activation functions in these aspects.
\subsection{Motivation and Contribution}
Previous activation functions can not satisfy all of the above five aspects. This motivates us to construct a novel function that satisfies these rules.

Our contributions include the following
\begin{itemize}
	\item  we present a novel activation function, which is odd, monotone, differentiable, has unbounded values and bounded continuous gradients. The function and its gradient can be efficiently evaluated.
	\item we analyze the properties of this function and argue why it is preferred as activation function.
	\item we numerically confirm that it performs better than others for many well-known neural networks.
\end{itemize}
\section{Truncated and Signed Square Root Function}
In this section, we present an activation function that fully satisfies the above five preferred properties. More specifically, we define an activation function as
\begin{equation}
	\label{eq:ours}
	f_{our}(x)=\left\{
	\begin{array}{ll}
		x, &\mathrm{when }\, |x|\le 1 \\
		\delta(x)(2\sqrt{|x|}-1), &\mathrm{else}
	\end{array}\right.,
\end{equation} where $\delta(x)$ is the sign of $x$ (if $x>0$, $\delta(x)=1$; if $x<0$, $\delta(x)=-1$). The square root function is truncated when $|x|\le 1$ because its gradient is dramatically increasing at $x=0$. Such truncation avoids the numerical issue. 

We name this function as Truncated and Signed Square Root (TSSR) function. This function and its gradient are illustrated in Fig.~\ref{fig1}, where we can tell the truncation indeed helps in avoiding the gradient issue at $x=0$.
\begin{table*}[!t]
	\caption{Comparison the activation functions\label{tab:table1}}
	\centering
	\begin{tabular}{c|c|c|c|c|c}
		\hline
		function & odd & monotone & differentiable & unbounded value range & continuous gradient\\
		\hline
		$f_1$ & $\usym{2717}$ & $\checkmark$ & $\checkmark$ & $\usym{2717}$& $\checkmark$ \\
		\hline
		$f_2$ & $\usym{2717}$ & $\checkmark$ & $\usym{2717}$ & $\usym{2717}$& $\usym{2717}$ \\
		\hline
		$f_3$ & $\usym{2717}$ & $\checkmark$ & $\usym{2717}$ & $\checkmark$& $\usym{2717}$ \\
		\hline
		$f_4$ & $\usym{2717}$ & $\checkmark$ & $\checkmark$ & $\usym{2717}$& $\usym{2717}$ \\
		\hline
		$f_5$ & $\usym{2717}$ & $\usym{2717}$ & $\checkmark$ & $\usym{2717}$& $\checkmark$ \\
		\hline
		$f_6$ & $\checkmark$ & $\checkmark$ & $\checkmark$ & $\usym{2717}$& $\checkmark$ \\
		\hline
		$f_7$ & $\checkmark$ & $\checkmark$ & $\checkmark$ & $\usym{2717}$& $\checkmark$ \\
		\hline
		$f_8$ & $\usym{2717}$ & $\usym{2717}$ & $\checkmark$ & $\usym{2717}$& $\checkmark$ \\
		\hline
		$f_9$ & $\usym{2717}$ & $\usym{2717}$ & $\checkmark$ & $\usym{2717}$& $\checkmark$ \\
		\hline
		$f_{10}$ & $\usym{2717}$ & $\usym{2717}$ & $\checkmark$ & $\usym{2717}$ & $\checkmark$ \\
		\hline
		$f_{our}$ & $\checkmark$ & $\checkmark$ & $\checkmark$ & $\checkmark$& $\checkmark$ \\
		\hline
	\end{tabular}
\end{table*}

The gradient of this signed and truncated function is
\begin{equation}
	\label{eq:ourg}
	f'_{our}(x)=\left\{
	\begin{array}{ll}
		1, &\mathrm{when }\, |x|\le 1 \\
		\frac{1 }{\sqrt{|x|}}, &\mathrm{else}
	\end{array}\right..
\end{equation} It follows that $0<f'_{our}(x)\le 1$. Therefore, the gradient never vanishes in the domain $(-\infty,+\infty)$. 

The gradient $f'_{our}(x)$ is continuous, showing its numerical stability. The continuity indicates that there is no dramatic change in a small neighborhood. In contrast, the gradient of $ReLU$ at $x_1=-0.001$ and $x_2=0.001$ are $\frac{\partial ReLU(x_1)}{\partial x}=0$ and $\frac{\partial ReLU(x_2)}{\partial x}=1$, respectively. This indicates that a small turbulence in the input might lead to a significant change in the gradient. Such discontinuity is harmful for the numerical stability of the neural networks.

The gradient $f'_{our}(x)$ is easy to implement in various programming languages such as Python, C++ and Java. Inverse of Square Root Function also has a fast numerical approximation because it frequently appears in computer graphics.

\subsection{Why Truncation?}
One might ask why not simply use the signed square root function directly. One reason is that the gradient of the signed square root function $\delta(x)\sqrt{|x|}$ has numerical issue at $x=0$. The truncation not only avoids this numerical issue at $x=0$ but also keeps the identity mapping in $[-1,1]$. 

The identity map around the origin avoids the model collapse from the activation function itself. It also makes the back-propagation more traceable.

\subsection{Mathematical Properties}
The proposed TSSR satisfies the preferred properties in the previous section.
\begin{itemize}
\item First of all, TSSR is odd, $f_{our}(-x)=-f_{our}(x)$. Such property guarantees that there is no bias from the activation function itself. And the zero point is mapped to itself $f(0)=0$. 
\item TSSR is increasing (monotone) because its gradient is always positive. Therefore, it is a bijective mapping. The monotone bijective mapping is important from information point of view. It means that there is no information collapsed or generated from the activation function itself. Moreover, the relative order from the input is also preserved in the output. 
\item TSSR is differentiable. Such smoothness guarantees that there is no dramatic change in a small neighborhood. 
\item TSSR has unbounded value range. This means that TSSR can still distinguish two input values even they are large. 
\item Its gradient is continuous and has bounded value range. Its nonzero gradient property guarantees that the vanishing gradient issue is not caused by the activation function itself, improving the networks' numerical stability. 
\end{itemize}

\subsection{Comparison with Others}
We compared the proposed TSSR with other activation functions. And the result is summarized in Table~\ref{tab:table1}. In the terms of the desired properties, TSSR can satisfy all of them, showing its advantages as activation function.

\begin{figure}
	\centering
	\includegraphics[width=0.4\linewidth]{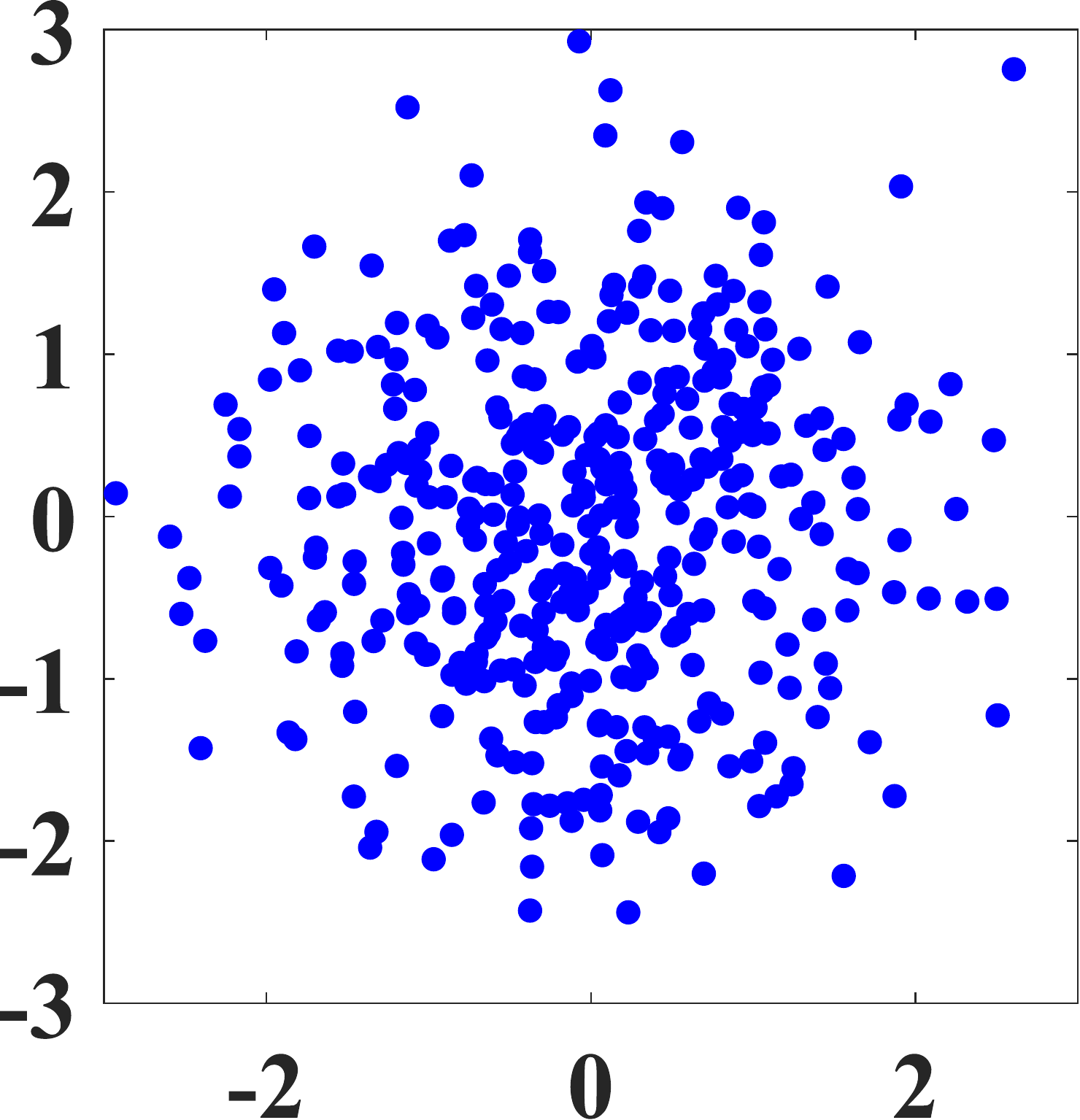}
		~~~
	\includegraphics[width=0.4\linewidth]{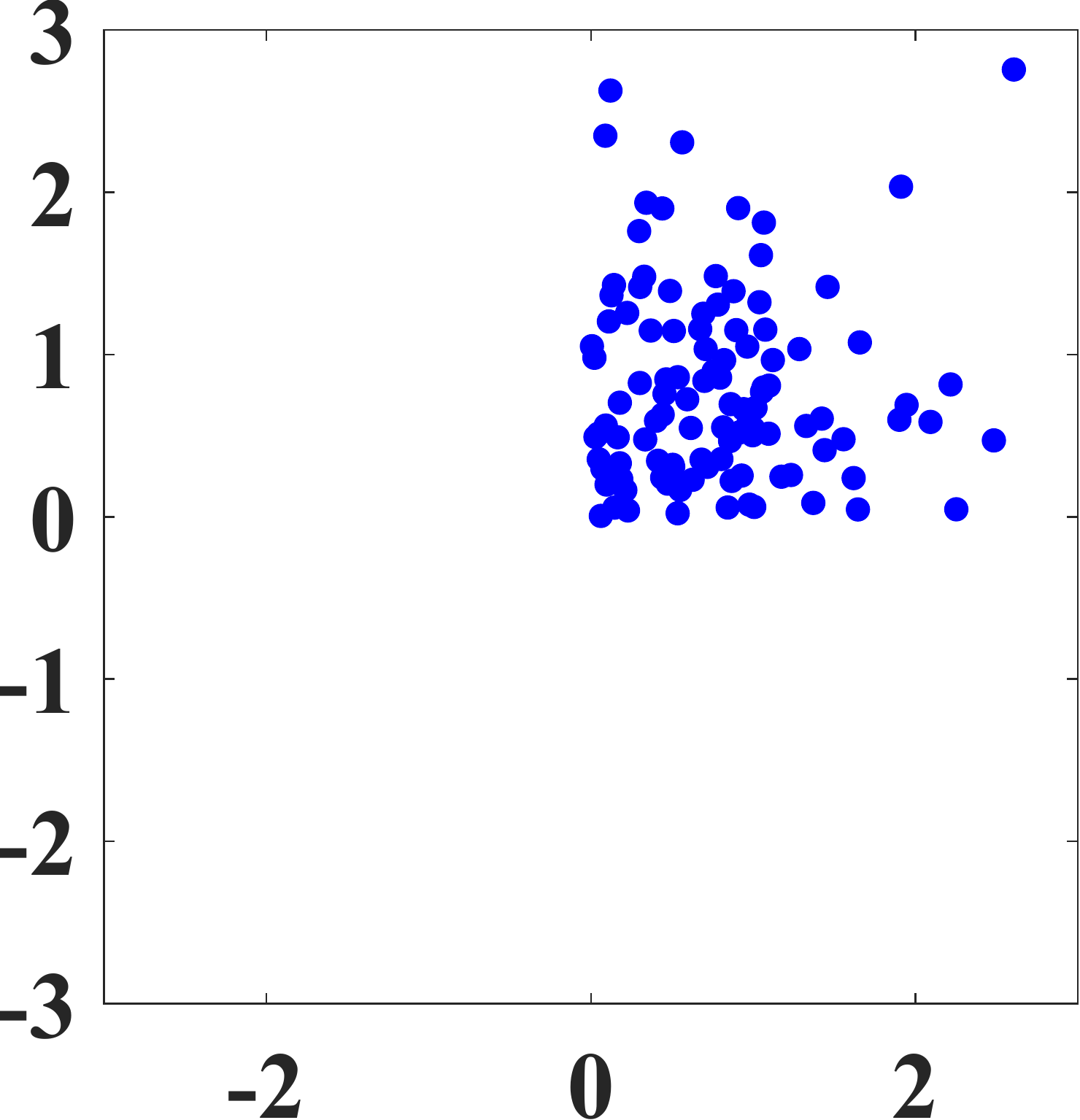}
	\textbf{input~~~~~~~~~~~~~~~~~~~~~~~~~~~~~ReLU}\\
	\includegraphics[width=0.4\linewidth]{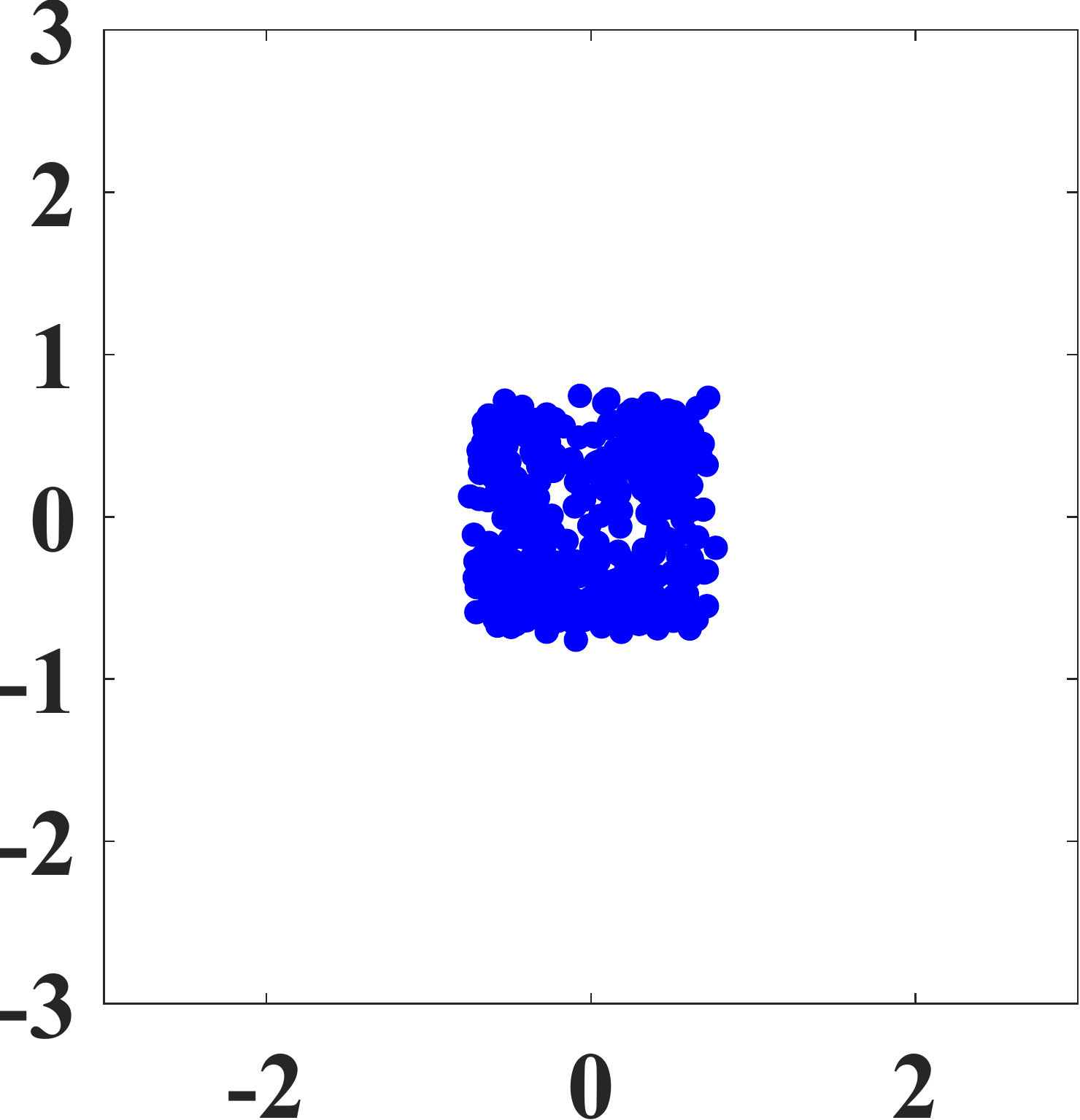}
	~~~
	\includegraphics[width=0.4\linewidth]{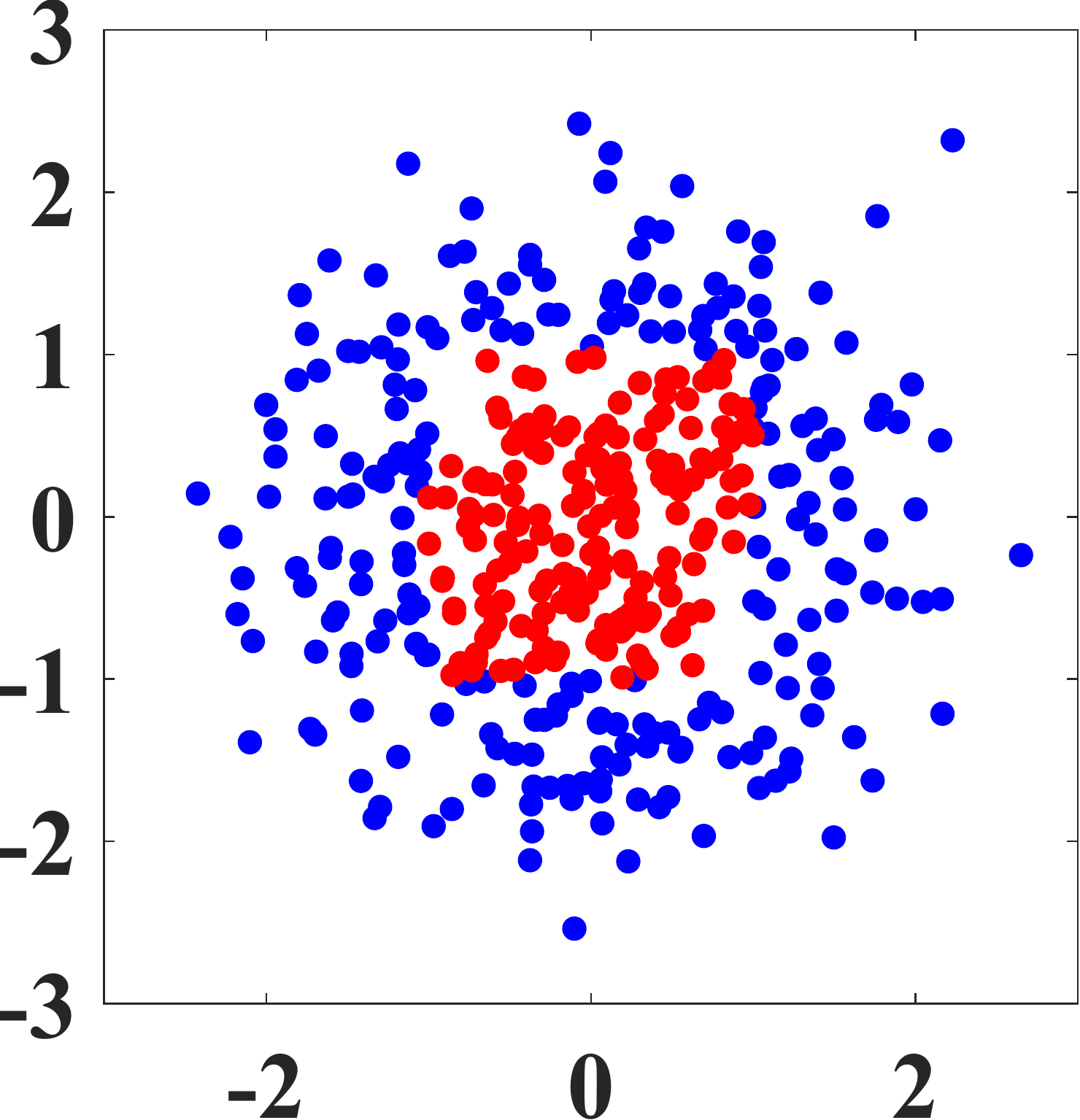}
	\textbf{Softsign~~~~~~~~~~~~~~~~~~~~~~~~~~~~~TSSR}\\
			\caption{The input is a point cloud satisfying normal distribution. After $ReLU$ activation function, the output only keeps the positive part. After $softsign$ activation function, the output are concentrated in the square region. After TSSR activation function, the input in the square region is preserved while other data is moving towards to the origin.}
			\label{figact}
		\end{figure}

A visual comparison between the proposed TSSR with two typical activation functions, $ReLU$ and $Softsign$ is shown in Fig.~\ref{figact}. The input is a point cloud that satisfies a normal distribution. After $ReLU$ activation function, the output only keeps the positive part. Therefore, all others are collapsed to zero. After $softsign$ activation function, the output are concentrated in the square region. Although this function is monotone, its bounded output range can not distinguish the inputs with large values. After TSSR activation function, the input in the square region (red) is preserved while other data (blue) is moving towards to the origin (nonlinear mapped).
\section{Experiments}
In this section, we numerically show the advantage of the proposed TSSR activation function in several well-known neural networks on the CIFAR dataset. We compare TSSR with ReLU~\cite{Nair2010}, Mish~\cite{Misra2020} and Serf~\cite{Nag2023}. In the future, we will compare TSSR with more activation functions in more neural networks on more datasets.

\subsection{CIFAR-10}
On CIFAR-10, we compared the proposed TSSR with ReLU, Mish and Serf in different neural networks, including SqueezeNet, Resnet-50, WideResnet-50-2, ShuffleNet-v2, ResNeXt-50, Inception-v3, DenseNet-121, MobileNet-v2, and EfficientNet-B0. In these networks, we only change the activation function. The top-1 \% accuracy values are shown in Table~\ref{tab:table2}. The results confirm that the proposed TSSR establishes a new state-of-the-art activation function. 
\begin{table}
	\caption{Comparison the activation functions in CIFAR-10 \label{tab:table2}}
	\centering
	\begin{tabular}{c|c|c|c|>{\columncolor{lightgray}}c}
		\hline
		Methods & ReLU &Mish &Serf & TSSR\\
		\hline
		SqueezeNet &84.14 &85.98 &86.32 & 86.63\\
		Resnet-50 & 86.54 &87.03 & 88.07 & 88.34\\
		WideResnet-50-2 &86.39 &86.57 &86.73 & 86.82\\
		ShuffleNet-v2 &83.93 &84.07 &84.55 & 84.78\\
		ResNeXt-50 & 87.25& 87.97 &88.49 &88.73\\
		Inception-v3 & 90.93 &91.55 &92.89 &92.91\\
		DenseNet-121 &88.59 &89.05& 89.07& 89.58\\
		MobileNet-v2 &85.74 &86.39& 86.61& 86.85\\
		EfficientNet-B0 (Swish)& 78.26 &78.02 &78.41 &78.86\\
		\hline
	\end{tabular}
\end{table}

\subsection{CIFAR-100}
On CIFAR-100, we compared the proposed TSSR with ReLU, Mish and Serf in different neural networks, including Resnet-164, WideResnet-28-10, DenseNet-40-12, and Inception-v3. In these networks, we only change the activation function. The top-1 \% accuracy values are shown in Table~\ref{tab:table3}. The results confirm that the proposed TSSR establishes a new state-of-the-art activation function.
\begin{table}[!t]
	\caption{Comparison the activation functions in CIFAR-100\label{tab:table3}}
	\centering
	\begin{tabular}{c|c|c|c|>{\columncolor{lightgray}}c}
		\hline
		Methods &ReLU &Mish &Serf &TSSR\\
		\hline
		Resnet-164 &74.55 &75.02 &75.13& 75.41\\
		WideResnet-28-10 &76.32 &77.03 &77.54 &77.62\\
		DenseNet-40-12 &73.68 &73.91 &74.16 &74.71\\
		Inception-v3 & 71.54 &72.38 &72.95&72.97\\
		\hline
	\end{tabular}
\end{table}

\section{Conclusion}
In this paper, we present a new activation function called the Truncated and Signed Square Root function (TSSR) for neural networks. The TSSR function has several advantages over other activation functions.

Firstly, the TSSR function is odd, which is a significant advantage compared to other activation functions. This means that it does not introduce any bias, ensuring that the neural network is not skewed towards one direction or another. This is particularly important in applications such as financial modeling or scientific simulations where accuracy is critical. Furthermore, the oddness of the TSSR function makes it more suitable for certain types of data. For example, data that is evenly distributed around the origin will benefit from the TSSR function's oddness, as it preserves the symmetry of the data.

Secondly, the function is monotonic, which is a key property for applications such as image recognition, classification, speech recognition, and natural language processing. It preserves the relative order of the input in the output, ensuring that the neural network is able to accurately distinguish between different inputs. This is particularly important in applications where precise ranking or ordering is required.

Thirdly, the function is differentiable, making it numerically stable and suitable for a wide range of applications, leading to faster and more accurate results. This differentiability also makes it easier to optimize the neural network during the training process. 

Fourthly, the function has an unbounded value range, which is particularly useful in applications where input values can vary widely. With the ability to distinguish between very large inputs, the TSSR function is well-suited for financial modeling and scientific simulations. 

Fifthly, the function has a continuous gradient, ensuring stable gradients during training. This leads to faster and more stable convergence in deep neural networks, which is essential for accurate results. The continuous gradient of the TSSR function also makes it easier to optimize and fine-tune the neural network during the training process.

Finally, the TSSR function and its gradient functions can be efficiently computed in programming languages, making it easy to implement the function in existing neural network frameworks. Its computational efficiency makes it an excellent choice for a wide range of applications, from image recognition to speech recognition. Additionally, the efficient computation of the TSSR function and its gradient functions make it more scalable, allowing for the use of larger datasets and more complex neural networks.

In conclusion, the TSSR function has significant advantages over other activation functions in neural networks. Its properties of oddness, monotonicity, differentiability, unbounded value range, continuous gradient, and computational efficiency make it an excellent choice for a wide range of applications. We believe that the TSSR function will play an important role in neural networks and machine learning, providing a new way to improve the accuracy and efficiency of these systems~\cite{chenouard:2014,gong2009symmetry,Lewis2019,zhao2023survey,Gong2012,Brown2020,gong2013a,Yu2019,Gong:2014a,Yin2019a,gong:phd,Yu2022a,gong:gdp,Guo2022,gong:cf,Zong2021,gong:Bernstein,Ezawa2023,Gong2017a,Tang2021a,Gong2018,Gong2018a,Yu2020,GONG2019329,Sancheti2022,Gong2019a,Tang2021,Gong2019,Yin2019b,Gong2022,Yin2020,Gong2020a,Jin2022,Gong2021,Tang2022,Gong2021a,Tang2022a,Tang2023,Gong2022,Tang2023a,Xu2023,Han2022,Gong2023a,Scheurer2023,Gong2023,Zhang2023b}.
\bibliographystyle{IEEEtran}
\bibliography{IEEEabrv,../../IP}

\vfill

\end{document}